\documentclass[conference]{IEEEtran}
\IEEEoverridecommandlockouts
\pdfoutput=1
\usepackage{cite}
\usepackage{amsmath,amssymb,amsfonts}
\usepackage{algorithmic}
\usepackage{graphicx}
\usepackage{textcomp}
\usepackage{xcolor}
\usepackage{bm}
\usepackage{multirow}

\def\BibTeX{{\rm B\kern-.05em{\sc i\kern-.025em b}\kern-.08em
    T\kern-.1667em\lower.7ex\hbox{E}\kern-.125emX}}
\begin{document}

\title{TraND: Transferable Neighborhood Discovery for Unsupervised Cross-domain Gait Recognition\\
}

\author{\IEEEauthorblockN{Jinkai Zheng\IEEEauthorrefmark{1}\thanks{This work is done when Jinkai Zheng is an intern at JD AI Research.},
Xinchen Liu\IEEEauthorrefmark{2},
Chenggang Yan\IEEEauthorrefmark{1},
Jiyong Zhang\IEEEauthorrefmark{1}\thanks{Corresponding author: Jiyong Zhang.},
Wu Liu\IEEEauthorrefmark{2},
Xiaoping Zhang\IEEEauthorrefmark{3} and
Tao Mei\IEEEauthorrefmark{2}}
\IEEEauthorblockA{\IEEEauthorrefmark{1}Automation School, Hangzhou Dianzi University, Hangzhou, China}
\IEEEauthorblockA{\IEEEauthorrefmark{2}AI Research of JD.com, Beijing, China}
\IEEEauthorblockA{\IEEEauthorrefmark{3}Ryerson University, Canada\\
\{zhengjinkai3, cgyan, jzhang\}@hdu.edu.cn, liuxinchen1@jd.com,
liuwu@live.cn,
xzhang@ee.ryerson.ca,
tmei@live.com\thanks{This work is supported by National Nature Science Foundation of China (61931008, 61671196, 62071415, 62001146, 61701149, 61801157, 61971268, 61901145, 61901150, 61972123), Zhejiang Province Nature Science Foundation of China (LR17F030006, Q19F010030), 111 Project, No. D17019.}}
}

\maketitle


\begin{abstract}
Gait, i.e., the movement pattern of human limbs during locomotion, is a promising biometric for identification of persons.
Despite significant improvement in gait recognition with deep learning, existing studies still neglect a more practical but challenging scenario --- unsupervised cross-domain gait recognition which aims to learn a model on a labeled dataset then adapt it to an unlabeled dataset.
Due to the domain shift and class gap, directly applying a model trained on one source dataset to other target datasets usually obtains very poor results.
Therefore, this paper proposes a Transferable Neighborhood Discovery (TraND) framework to bridge the domain gap for unsupervised cross-domain gait recognition.
To learn effective prior knowledge for gait representation, we first adopt a backbone network pre-trained on the labeled source data in a supervised manner.
Then we design an end-to-end trainable approach to automatically discover the confident neighborhoods of unlabeled samples in the latent space.
During training, the class consistency indicator is adopted to select confident neighborhoods of samples based on their entropy measurements.
Moreover, we explore a high-entropy-first neighbor selection strategy, which can effectively transfer prior knowledge to the target domain.
Our method achieve the state-of-the-art results on two public datasets, i.e., CASIA-B and OU-LP.
\end{abstract}

\begin{IEEEkeywords}
Gait Recognition, Human Identification, Domain Adaptation, Neighborhood Discovery, Deep Learning
\end{IEEEkeywords}

\section{Introduction}
Visual gait recognition aims to identify a person using the gait sequences captured by multiple cameras.
This task has been studied for over a decade since gait is a discriminative biometric that can be remotely obtained without the cooperation of subjects~\cite{csur/WanWP19}.
Due to the significant success of deep learning for various computer vision tasks~\cite{eccv/LiuLMM16, tmm/LiuLMM18, tmm/YanXCZHZD18, tmm/YanLZLZD19, tmm/YanTWZHZD20, tomm/abs-2008-03741, tmm/YanSZNZX20, tpami/abs-2002-00169}, deep learning based methods for gait recognition have also achieved excellent performance on individual datasets~\cite{pami/WuHWWT17,  cvpr/ZhangL00L19}.
However, in real-world applications, it is more practical to learn a model on a dataset collected from one scene (i.e., the source domain) while applying it to another scene (i.e., the target domain).
Moreover, one may have only labels of the source domain but no labels of the target data since it is usually difficult and expensive for large-scale annotation.
Therefore, this paper concentrates on the unsupervised domain adaptation problem for gait cognition, which is a valuable task yet overlooked by the community of computer vision and pattern recognition.

\begin{figure}[t]
	\centering
	{\includegraphics[width=\columnwidth]{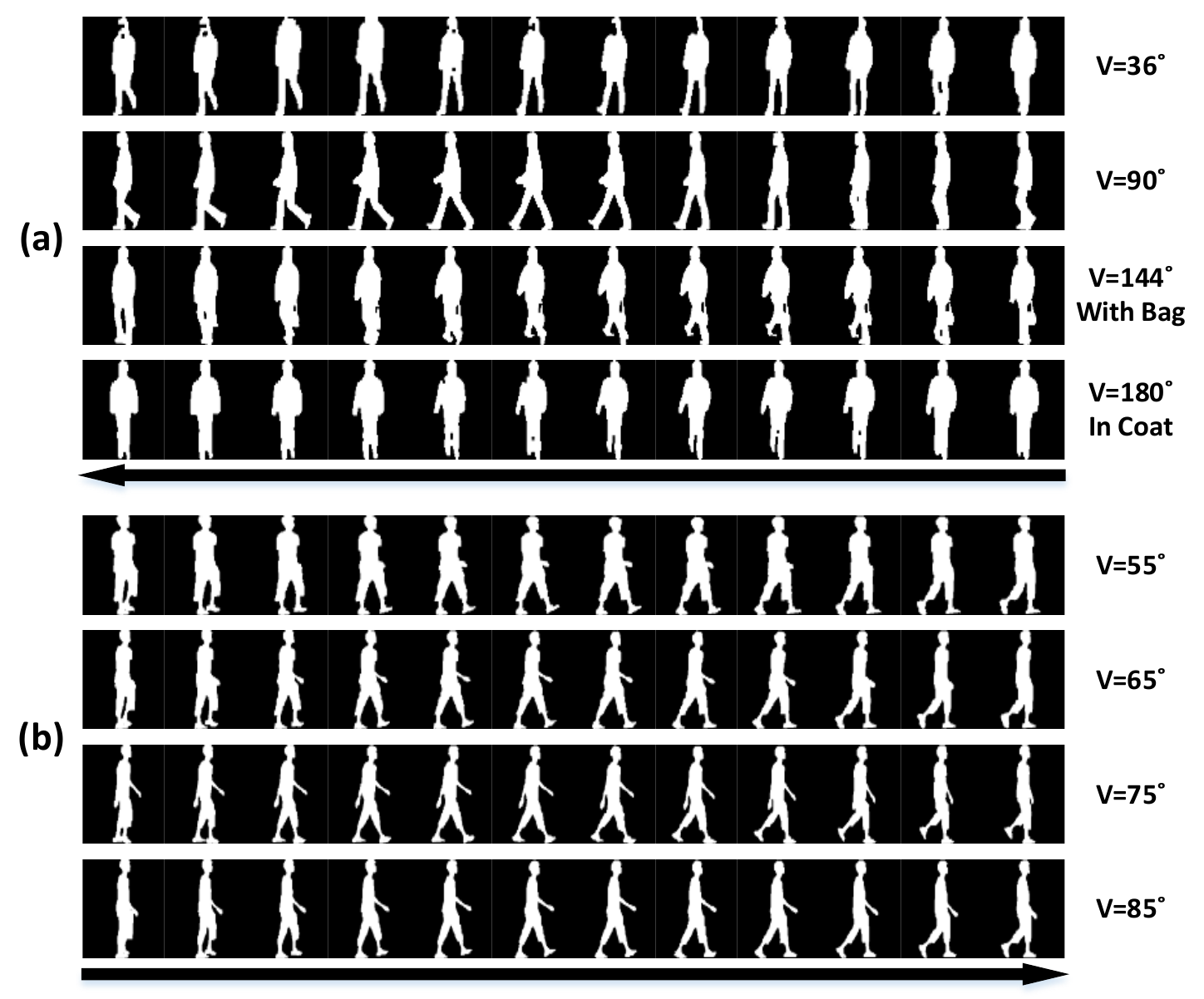}}
	\caption{(a) Gait sequences of one person in the CASIA-B dataset in different conditions and viewpoints. (b) Gait sequences of one person captured from four viewpoints in the OU-LP dataset.}
	\label{fig:figure1} \vspace{-5mm}
\end{figure}

However, unsupervised domain adaptation for gait recognition is a non-trivial task that faces several challenges as shown in Fig.~\ref{fig:figure1}.
First of all, there is a large domain shift between the source and the target datasets.
Similar to domain adaptation for image classification, one of the main challenges is the variance in the spatial dimension of images~\cite{ijon/WangD18}.
Varied camera settings, clothing conditions, and scenes make the visual style of gait sequences very different across different datasets.
Moreover, compared with recognition tasks in still images, the domain shift for cross-domain gait recognition also results from the temporal dimension, such as the varied walking rhythms and habits in different scenes or nations~\cite{tifs/IwamaOMY12}.
Furthermore, another challenge is the class gap between datasets, because the classes/IDs of persons have no overlap across any two datasets~\cite{cvpr/YouLCWJ19}.
It is more difficult under the unsupervised condition for which we have only the labels for the source dataset but no annotation for the target dataset.

Vision-based gait recognition has been studied for over ten years~\cite{tip/MuramatsuSMUY15, csur/WanWP19}.
From hand-crafted features~\cite{pami/HanB06, iccv/LombardiNMY13} to data-driven representations learned by deep Convolutional Neural Network (CNN)~\cite{icb/ShiragaMMEY16, cvpr/MakiharaSMLY17, aaai/ChaoHZF19} or Recurrent Neural Network (RNN)~\cite{cvpr/ZhangT0A0WW19, tmm/LiLM19}, the accuracy of existing methods has been greatly improved.
However, they only consider learning and testing models on individual datasets while neglecting the cross-dataset condition.
Researchers have studied the unsupervised cross-domain person re-identification (Re-Id) of which classes of persons also have no overlap across different datasets~\cite{cvpr/PengXWPGHT16, cvpr/Deng0YK0J18, cvpr/Zhong0LL019}.
They usually first adopt a CNN pre-trained on labeled source data to extract features from unlabeled target data.
Then a clustering approach, i.e., k-means~\cite{tomccap/FanZYY18} or DBSCAN~\cite{iccv/YangYHXHT19}, is applied to assign pseudo labels for the target training set. 
At last, the CNN is fine-tuned using target samples with pseudo labels.
However, since the number of IDs in the target domain is unknown, it is hard to set suitable hyper-parameters for clustering methods, such as the center number of k-means.
Moreover, the model cannot obtain optimal performance due to the noise in pseudo labels. 

To this end, we propose a Transferable Neighborhood Discovery framework, named as TraND, for unsupervised cross-domain gait recognition.
To learn effective features from gait sequences, we adopt GaitSet~\cite{aaai/ChaoHZF19} as the backbone in TraND, since it is the state-of-the-art model to capture discriminative and robust features for gait recognition.
We first train the GaitSet network on the source dataset with labels to learn prior knowledge for gait representation.
Different from existing methods~\cite{tomccap/FanZYY18, iccv/YangYHXHT19}, we adopt an end-to-end trainable approach to automatically discover the neighborhoods of unlabeled samples inspired by the anchor neighborhood discovery (AND) for unsupervised image recognition~\cite{icml/HuangDGZ19}.
During training, the class consistency indicator is adopted to select confident anchors and their neighbors based on their entropy distribution.
In contrast to original AND, we explore a high-entropy-first sample selection strategy, which can effectively transfer prior knowledge to the target domain.
At last, with confident neighborhoods, the model is optimized with the anchor neighborhood loss.

In summary, the contributions of this paper include:
1) we make one of the first attempts for unsupervised cross-domain gait recognition which is a valuable yet unexplored task;
2) we design a TraND framework to learn the transferable representation for gait recognition, which can close the domain gap between datasets by automatic neighborhood discovery;
3) our method achieves state-of-the-art results across two large-scale datasets, i.e., CASIA-B and OU-LP, which show the effectiveness of our method.

\begin{figure*}[t]
	\centering
	{\includegraphics[width=0.9\textwidth]{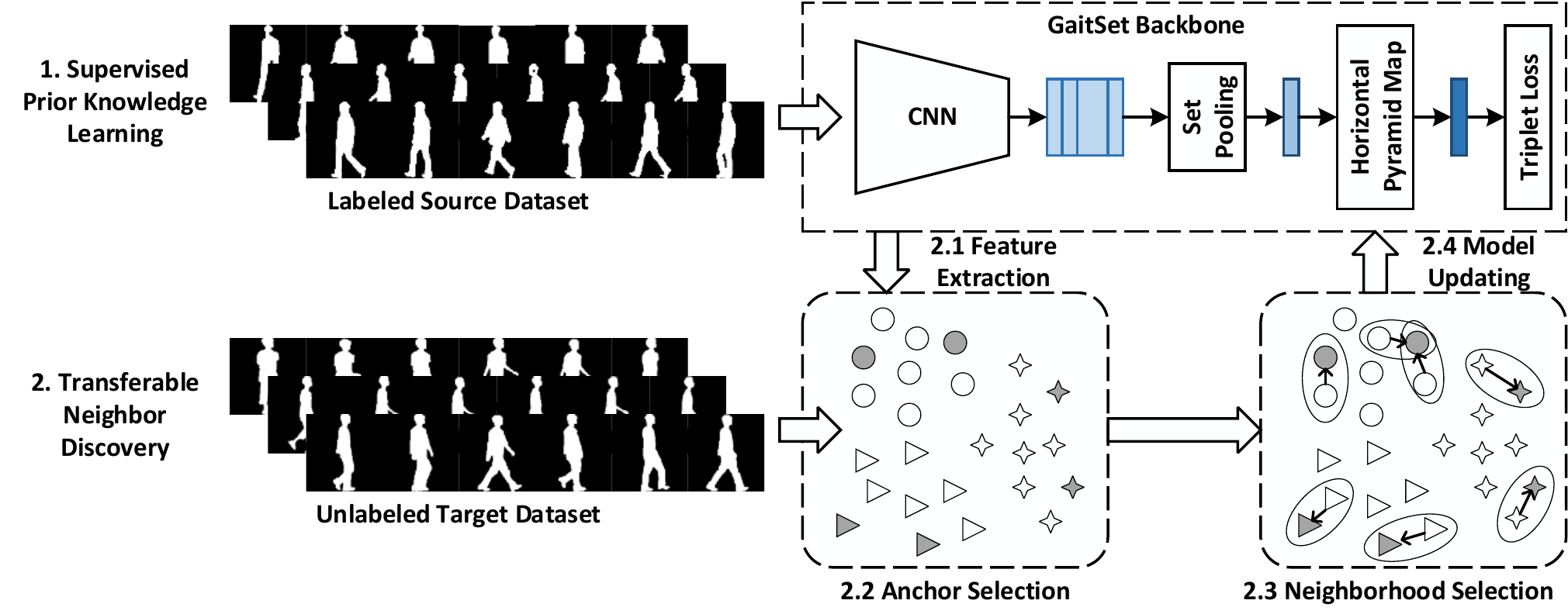}}
	\caption{The overall framework of the Transferable Neighbor Discovery framework.}
	\label{fig:figure2}\vspace{-5mm}
\end{figure*}

%

\section{The Proposed Framework}

\subsection{Overview}
Fig.~\ref{fig:figure2} shows the structure of the Transferable Neighborhood Discovery method with two main stages.
In the first stage, we adopt GaitSet~\cite{aaai/ChaoHZF19} as the backbone network to learn gait features from silhouette sequences.
We train the GaitSet network on the labeled source dataset to learn the prior knowledge of gaits.
In the second stage, we first map unlabeled target samples into the feature space with the trained backbone.
After that, target samples are distributed in a manifold with the prior knowledge learned from the source data.
Then we adopt the class consistency indicator measured by the entropy of samples to select the confident neighborhoods of samples in a progressive manner inspired by~\cite{icml/HuangDGZ19}.
We explore a high-entropy-first strategy to select confident neighborhoods, which are finally used to update the network.

\subsection{GaitSet as the Backbone}
\label{sec:backbone}
To learn discriminative representation from gait sequences, we adopt the state-of-the-art CNN based model, i.e., GaitSet~\cite{aaai/ChaoHZF19}, as the backbone network in the TraND framework, as shown in Fig.~\ref{fig:figure2}.
The GaitSet network directly takes the silhouette sequence as the input, which considers the sequence as a set but ignores the order of frames based on the assumption that the
appearance of a silhouette has contained its position information.
By this means, the spatial features of all frames can first be kept to comprehensively model the gait representation.
We then aggregate the frame-level features into a sequence-level feature with the Set Pooling (SP) operation as in GaitSet~\cite{aaai/ChaoHZF19}.
At last, to discover the multi-scale features, the Horizontal Pyramid Mapping (HPM) is applied to generate a discriminative representation of gait sequences, which splits the feature map extracted by SP into $\sum_{s=1}^{5} 2^{s-1}$ strips on height dimension.
Given a sequence of gait silhouettes, $X = \{x_k\}^K$, where $x_k$ is one silhouette image and $K$ is the length of the sequence, the overall process of the GaitSet network, $F(\cdot)$, can be formulated as follows:
\begin{equation}
\label{equ1}
F(X) = h(g(\{f(x_k)\}^K)),
\end{equation}
where $f(\cdot)$ is the CNN to extract the frame-level feature, $g(\cdot)$ is the SP function to generate the sequence-level feature, and $h(\cdot)$ is the HPM operation as in GaitSet~\cite{aaai/ChaoHZF19}.

\subsection{Supervised Learning of Prior Knowledge}
\label{sec:supervised}
Since we only have labels of the source data but no knowledge about the target domain, it is necessary to effectively exploit the prior knowledge for gait representation in the source dataset.
Inspired by methods of unsupervised person Re-Id~\cite{cvpr/Zhong0LL019}, we train the backbone on the source dataset $\{ \mathcal{X}_S, \mathcal{Y}_S \}$ as shown in the upper part of Fig.~\ref{fig:figure2}.
Since the IDs in the source data and the target data are not overlapped, it is better to make the model learn the similarity or difference between samples rather than classify them.
So, we adopt the Triplet Loss~\cite{cvpr/SchroffKP15} instead of Cross-Entropy Loss for prior knowledge learning.
The formulation of the Triplet Loss is
\begin{equation}
\label{equ2}
L = \sum_{i=1}^{N} [||F(X_i^a)-F(X_i^p)|| - ||F(X_i^a)-F(X_i^n)|| + m]_{+},
\end{equation}
where $N$ is the number of training sequences, $X_i^a$ is the anchor, $X_i^p$ and $X_i^n$ are positive and negative sequences with respect to the anchor, respectively, and $m$ is the margin parameter.
Here, all samples are randomly selected from the source dataset $\mathcal{X}_S$.

\subsection{Knowledge Transfer by Neighborhood Discovery}
\vspace{-2mm}
\label{sec:USTA}
Base on the prior knowledge, the trained backbone network can support a discriminative latent space that makes the samples of different persons in the source domain differentiable.
Therefore, we can map the samples of the target dataset into this feature space by feeding them into the backbone to obtain their primary representations in the feature space.
However, due to the domain gap between the source domain and the target domain, the samples in the target domain can be shifted that the gait sequences of the same person may be far from each other while the sequences of different persons may be close.
If we directly apply the clustering method on the target features based on their pairwise distances, the ambiguous samples can bring noises for refining the model.
Moreover, since the number of classes/IDs in the target dataset is unknown, it is difficult to set proper parameters for the clustering methods, such as $K$ in the k-means algorithm.

Inspired by the AND method for unsupervised deep learning~\cite{icml/HuangDGZ19}, we present the TraND framework to transfer the knowledge from the source data to the target data by neighborhood discovery.
Here we introduce how to discover neighborhoods with the learned prior knowledge, as shown in the lower part of Fig.~\ref{fig:figure2}.
First of all, we define the anchor neighborhood (AN) using $k$ nearest neighbors ($k$NN) as in~\cite{icml/HuangDGZ19}.
Given a sample of gait sequence $X_t \in \mathcal{X}_T$, a neighborhood $\mathcal{N}_k(X_t)$ anchored to $X_t$ is formulated as:
\begin{equation}
\label{equ3}
\mathcal{N}_k(X_t) = \{X_i \ | \ s(X_i, X_t) {\rm \ is \ the \ top-}k {\rm \ in \ } S\} \cup \{X_t\},
\vspace{-2mm}
\end{equation}
where $S$ is the feature space, $s(\cdot, \cdot)$ is a pairwise similarity metric such as the cosine similarity.

\textbf{Neighborhood Selection with Prior Knowledge.}
Due to the domain gap between the source dataset and the target dataset, it is hard to select confident samples with class-consistent neighbors.
Inspired by AND~\cite{icml/HuangDGZ19}, we adopt the class consistency indicator $H(X_t)$, where the larger the value of $H(X_t)$ is, the more reliable of the corresponding sample is.
Given a gait sequence $X_i \in \mathcal{X}_T$, $H(X_i)$ is formulated as:
\vspace{-2mm}
\begin{equation}
\label{equ4}
H(X_i) = - \sum_{j=1}^{N}p_{i,j}log(p_{i,j}), \ p_{i,j} = \frac{exp(X_i^{\top} X_j /\tau)}{\sum_{k=1}^{N} exp(X_i^{\top} X_k /\tau)},
\end{equation}
where $p_{i,j}$ is the non-parametric softmax measurement and $\tau$ is a temperature for controlling distribution concentration.

The $p_{i,j}$ represents the degree of the similarity between $X_i$ and $X_j$, and the larger $p_{i,j}$ is, the more similar $X_i$ is to $X_j$.
Since AND~\cite{icml/HuangDGZ19} is designed for unsupervised learning without any prior knowledge on the training data, the samples in the feature space distribution like the instance discrimination~\cite{cvpr/WuXYL18}, only a few samples get high similarity, which means that these samples are far away from other samples, i.e., these samples will be far away from the large group and reside in a low-density area with sparse visual similar neighbours surrounding. 
After the operation of softmax, the $p_{i,j}$ will be large and the corresponding $H(X_i)$ will be small, which means $X_i$ resides in a low-density area. Therefore, in the original AND, the low-entropy-first strategy is exploited to select the most reliable samples for optimization.
However, in our TraND, more visually similar samples of the target domain have latent relationships due to the prior knowledge learned from the source data, These samples that reside in dense area has small $p_{i,j}$ and the corresponding $H(X_i)$ will be large. In order to better use the knowledge learned from the source domain, here we adopt the high-entropy-first strategy  for neighborhood selection which is proved the effectiveness by the further experiment results.

\begin{table*}[t]
	\centering
	\caption{Rank-1 (Rank-1 excluding identical views in brackets) results across the CASIA-B (CA) and OU-LP (OU) datasets.}\vspace{-3mm}
	\label{tab:table1}
	\begin{tabular}{l|c|c|c|c}\hline
		\multirow{2}{*}{Method}			& \multicolumn{3}{c|}{OU$\Rightarrow$CA}   	& {CA$\Rightarrow$OU} 	\\ \cline{2-5}
		& {NM}   			& {BG}   		& {CL}   		&  {NM}   					\\ \hline
		Supervised GaitSet~\cite{aaai/ChaoHZF19}	& 95.5 (95.1)		& 88.8 (87.8)	& 71.5 (69.8) 	& 99.2 (99.1)  			\\
		Direct Testing 	    						& 45.9 (40.6)		& 34.7 (29.7)	& 13.3 (11.3)	& 69.0 (60.2) 			\\ \hline
		GaitSet + DBSCAN~\cite{iccv/YangYHXHT19}	& 51.3 (46.5) 		& 39.7 (34.9)	& 17.4 (15.2) 	& 64.5 (53.6)  			\\
		GaitSet + k-means~\cite{tomccap/FanZYY18}	& 44.2 (38.6) 		& 36.2 (30.7)	& 16.1 (14.7) 	& 58.5 (46.4)  	\\ 
		GaitSet + AND~\cite{icml/HuangDGZ19}		& 45.9 (40.6)		& 28.7 (24.4) 	& 13.9 (12.3) 	& 50.2 (41.0)  			\\ \hline
		TraND (Random Entropy)        				& 41.5 (35.9)		& 27.8 (23.7) 	& 11.1 (9.7) 	& 52.5 (41.1)  			\\
		TraND (Low Entropy)					& 46.3 (41.1) 		& 32.5 (27.7)	& 11.2 (9.7)	& 69.1 (61.0)  			\\
		TraND (Ours)                        				& \textbf{66.7 (63.4)}& \textbf{46.5 (42.7)}	& \textbf{17.5 (15.8)}	& \textbf{80.0 (75.6)	}	\\ \hline
	\end{tabular}\vspace{-5mm}
\end{table*}



\textbf{Model Updating with Neighborhood Supervision.}
Inspired by curriculum learning~\cite{icml/BengioLCW09}, we refine the backbone CNN in a from-easy-to-hard manner, by which we train the backbone in $R$ rounds.
For each round $r$, we apply above the neighborhood selection strategy to obtain the top $r/R \times 100 \% $ anchor samples with their neighbors to update the backbone.
At last, the backbone is optimized with the unsupervised anchor neighborhood loss which is formulated as:
\begin{equation}
\label{equ5}
\mathcal{L}_{AN} = - \sum_{i=1}^{N_r}log(\sum_{j \in \mathcal{N}_k(X_i)} p_{i,j}).
\end{equation}

\section{Experiments}

\subsection{Experimental Setting}
\textbf{Datasets.}
The experiments is performed on two large-scale gait datasets, i.e., CASIA-B~\cite{icpr/YuTT06} and OU-LP~\cite{tifs/IwamaOMY12}.
The CASIA-B dataset contains 124 persons captured from 11 viewpoints ($0^\circ$, $18^\circ$, ..., $180^\circ$).
Each person walks six times under normal conditions (NM \#1-6), two times in their coats (CL \#1-2), and two times with bags (BG \#1-2) to obtain ten gait sequences in total.
Overall, the CASIA-B dataset contains 13,640 sequences.
Following~\cite{tmm/LiLM19, aaai/ChaoHZF19}, 74 persons are used for training and the rest 50 persons are leaved for test. 
For each subject of the testing set, the first four normal sequences are registered as the gallery while the rest sequences are probes.
The OU-LP dataset has 4,016 subjects of which 3,836 subjects are adopted for the cross-camera person identification task.
For each subject, two gait sequences are captured by cameras and split into four subsequences based on viewpoints to the cameras (i.e., $55^\circ$, $65^\circ$, $75^\circ$, and $85^\circ$).
In our experiment, the 60\% of the 3,836 subjects are used for training and the rest are for testing.
For each subject in the testing set, the first sequence is registered as the probe and the other is the gallery.


\textbf{Evaluation Metric.}
The evaluation is performed in a cross-dataset manner.
So, the model is trained with the training set of CASIA-B with labels and that of OU-LP without labels while is evaluated on the testing set of OU-LP, and vice versa.
During testing, we compute the Euclidean distance between each probe and all gallery, which is similar to person Re-Id~\cite{iccv/ZhengSTWWT15}.
The rank-1 accuracy for each probe is calculated, while the overall result is measured by the average rank-1 accuracy over all probes.

\textbf{Implementation Details.}
In our experiments, we adopt GaitSet as the backbone to be trained on the source dataset using settings in~\cite{aaai/ChaoHZF19} and~\cite{corr/abs-2006-02631}.
The margin in Equ.~\ref{equ2} is set as $0.2$.
Especially, a batch with size of $p \times k$ is sampled from the training set where $p$ denotes the number of persons and $k$ denotes the number of training samples for each person.
We set the batch size as $8 \times 16$ and $16 \times 8$ for CASIA-B and OULP, respectively.
For the unsupervised learning on the target dataset, the number of nearest neighbors, K, in Equ.~\ref{equ3} is set as 1.
The $\tau$ of $p_{i,j}$ in Equ.~\ref{equ5} is set as 0.1.
During model updating with neighborhood discovery, we train the backbone in $R=4$ rounds and 200 epochs per round.
The initial learning rate is set to $10^{-5}$ and scaled-down by 0.1 in every 40 epochs after the 80-th epoch.
The experiments were performed on four Tesla P40 GPUs.

\subsection{Comparison with the state-of-the-art methods}
We compare TraND with seven methods:
1) \textbf{Supervised GaitSet}~\cite{aaai/ChaoHZF19}: the GaitSet trained on target dataset in the supervised manner as the upper bound;
2) \textbf{Direct Testing}: directly applying the model trained on the source dataset to the target testing set;
3) GaitSet with clustering methods: \textbf{GaitSet + DBSCAN}~\cite{iccv/YangYHXHT19} and \textbf{GaitSet + k-means}~\cite{tomccap/FanZYY18};
4) \textbf{GaitSet + AND}: directly using AND to train the target domain in an unsupervised method with Gaitset as the backbone~\cite{icml/HuangDGZ19};
5) Variants of TraND: \textbf{TraND (Random entropy)} and \textbf{TraND (Low entropy)} which select neighborhoods with random entropy and low entropy, respectively.

The experimental results are listed in Table~\ref{tab:table1}.
From the results, we can first find that directly applying the model trained on the source data to the target data obtains very poor results compared with the model trained on the target data.
It shows that the domain gap among different datasets can greatly affect the adaptation capability of the model.
Moreover, the clustering-based methods, i.e., GaitSet + DBSCAN and GaitSet + k-means, cannot work well for this task.
This is because the pseudo labels generated by clustering algorithms may be misleading, which brings much noise during fine-tuning the model.
Furthermore, by comparing different neighborhood selection strategy, our method achieves the best performance.
This reflects that our method can effectively discover confident neighborhoods with the class consistency indicator and high-entropy-first strategy.
However, when transferring the model trained on OU-LP to the conditions of BG and CL on CASIA-B, all methods obtain poor results.
This is because OU-LP has no samples with a bag or coat, which makes models learn nothing about this knowledge.
Therefore, the cross-condition settings should be further studied to improve the performance.

\section{Conclusion}
In this paper, we make one of the first explorations for unsupervised cross-domain gait recognition with a TraND framework.
In TraND, we first adopt a strong backbone network to learn the prior knowledge from the labeled source data.
Then the unlabeled target samples are mapped into the feature space supported by the trained backbone.
We adopt the class consistency indicator and a high-entropy-first strategy to progressively select confident neighborhoods, which are utilized to optimize the model using the anchor neighborhood loss.
At last, the superior results of TraND across two public datasets show the effectiveness of our method.

\bibliographystyle{plain}
\bibliography{iscas2021abbr}

\end{document}